\documentclass[letterpaper, 10 pt, conference]{ieeeconf}

\IEEEoverridecommandlockouts
\usepackage{amsmath,amsfonts}
\usepackage{algorithmic}
\usepackage{algorithm}
\usepackage{array}
\usepackage[caption=false,font=normalsize,labelfont=small,textfont=small]{subfig}
\usepackage{textcomp}
\usepackage{stfloats}
\usepackage{url}
\usepackage{verbatim}
\usepackage{graphicx}
\usepackage{cite}
\usepackage{color}
\usepackage{soul}
\usepackage{pifont}
\usepackage{overpic}
\usepackage{booktabs}
\usepackage{multirow}
\usepackage{siunitx}
\usepackage[table,xcdraw]{xcolor}
\usepackage[capitalise, nameinlink]{cleveref}
\usepackage{bm}

\soulregister\cite7
\begin{document}

\title{\LARGE \bf Tendon-driven Grasper Design for Aerial Robot \\ Perching on Tree Branches}

\author{Haichuan Li$^{1}$, Ziang Zhao$^{1}$, Ziniu Wu$^{1}$, Parth Potdar$^{2}$, Long Tran$^{3}$, \\ Ali Tahir Karasahin$^{1,4}$, Shane Windsor$^{1}$, Stephen G. Burrow$^{1}$ and Basaran Bahadir Kocer$^{1}$
% \author{IEEE Publication Technology,~\IEEEmembership{Staff,~IEEE,}
        % <-this % stops a space
\thanks{$^{1}$Authors are with the School of Civil, Aerospace and Design Engineering,
University of Bristol, \texttt{b.kocer@bristol.ac.uk}. }
\thanks{$^{2}$Parth Potdar is with the Department of Engineering, University of Cambridge.}
\thanks{$^{3}$Long Tran is with the School of Engineering Mathematics and Technology,
University of Bristol.}
\thanks{$^{4}$Ali Tahir Karasahin is with Faculty of Engineering, Department of Mechatronics Engineering, Necmettin Erbakan University, Turkey.}%
}% <-this % stops a space
%\thanks{Manuscript received X, X; revised X, X.}}

% \IEEEpubid{0000--0000/00\$00.00~\copyright~2021 IEEE}
% Remember, if you use this you must call \IEEEpubidadjcol in the second
% column for its text to clear the IEEEpubid mark.

\maketitle

\begin{abstract}
Protecting and restoring forest ecosystems has become an important conservation issue. Although various robots have been used for field data collection to protect forest ecosystems, the complex terrain and dense canopy make the data collection less efficient. To address this challenge, an aerial platform with bio-inspired behaviour facilitated by a bio-inspired mechanism is proposed. The platform spends minimum energy during data collection by perching on tree branches. A raptor inspired vision algorithm is used to locate a tree trunk, and then a horizontal branch on which the platform can perch is identified. A tendon-driven mechanism inspired by bat claws which requires energy only for actuation, secures the platform onto the branch using the mechanism's passive compliance. Experimental results show that the mechanism can perform perching on branches ranging from 30 \si{\milli \metre} to 80 \si{\milli \metre} in diameter. The real-world tests validated the system's ability to select and adapt to target points, and it is expected to be useful in complex forest ecosystems. Project website: \url{https://aerialroboticsgroup.github.io/branch-perching-project/}

\end{abstract}

% \begin{IEEEkeywords}
% Aerial Robotics, Target Recognition, Perching Mechanism Design
% \end{IEEEkeywords}

%%%%%%%%%%%%%%%%%%%%%%%%%%%%%%%%%%%%%%%%%%%%%%%%%%%%
\section{Introduction}
%%%%%%%%%%%%%%%%%%%%%%%%%%%%%%%%%%%%%%%%%%%%%%%%%%%%
Aerial robotic systems play an important role in many forest management scenarios such as image acquisition \cite{ho2022vision,bates2025leaf}, sample collection \cite{kocer2021forest} and environmental exploration \cite{lan2024aerial}. The miniaturisation and commercialisation of onboard sensors have enabled aerial robots to carry out intelligent task execution, resulting in significant savings in manpower and material resources in multiple tasks. In order to improve the endurance of aerial robots for long-term observation and sample collection in field environments, numerous studies have focused on the perching of aerial robots \cite{hauf2023learning}.

\par Many researchers have focused on drone perching on planar surfaces, using magnets or similar devices to assist drones in perching on stationary or moving planar surfaces, such as \cite{gao2023adaptive,do2023vision,mao2023robust}. These works have empowered drones to perch and dock with both efficiency and precision by utilising onboard cameras for visual servo control and corresponding aerial robotics control algorithms. However, these perching method targets regular surfaces as their perching sites, making it difficult to realise the perching task on complex, irregular surfaces in the field. 
In aerial robot field perching missions, identifying the characteristics of the perched target becomes critical to achieve a successful perching. 
A growing number of researchers are working to extend planar perching for aerial robots in relatively ideal environments to more general, environmentally characterised scenarios. In \cite{kominami2023detection}, a depth camera and position estimation fusion method are presented and combined with a simple gripping mechanism that assists an aerial robot in landing on a 2 \si{\milli\metre} thin board. And a modular perching method capable of recognising different perching positions such as object edges, sharp corners or performing grasping actions proposed in \cite{hang2019perching}, endowing aerial robots with higher flexibility and adaptability during perching. These studies have struggled with perching point selection in cluttered forest environments due to shape limitations in visual discrimination.

\begin{figure}[t]
		% Requires \usepackage{graphicx} 
		\centering
		\begin{overpic}[scale=0.33,trim=120 0 100 50,clip]{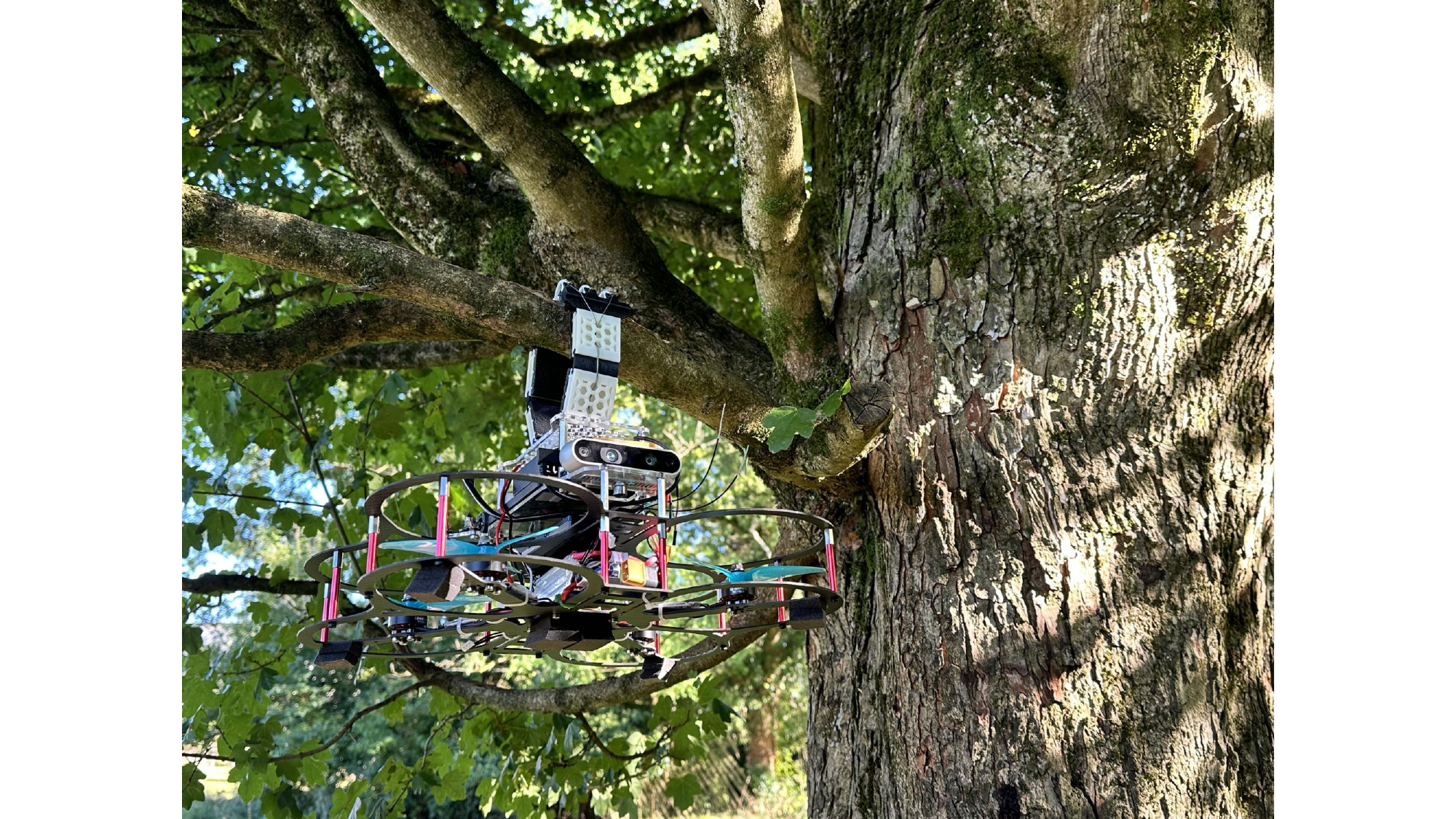}
			%\put(1,70){\text{E(J)}}
			%\put(38,55){\text{310mm}}
			%\put(77,43){\text{278mm}}
			%\put(78,69){\text{10cm}}
			%\put(86,3){\text{D(mm)}}
			%\put(15,3){Stable state 
		\end{overpic}
\caption{An example of aerial robotic platforms inhabiting real natural environments.}\label{fig1}
\end{figure} 

\par To further overcome the limitations of specific surfaces or shapes, researchers have further sought solutions for acquiring the contours of trees in the forest for aerial robot perching. In \cite{bai2024design}, a bat-like drone perching mechanism is proposed and combined with a depth camera, a single-point LiDAR, and a pressure sensor for aerial robot tracking and autonomous perching of target points. In addition, many studies such as \cite{busch2021dynamic,lin2024drone,tong2023image} have investigated the operation and interaction of aerial robots in unstructured environments, aiming to extract the environmental characteristics suitable for drone operation, which can guide the design of drone control and mechanism, and is expected to further expand the ability of aerial robots to perch in complex forest environments.
\par To improve the recognition capability of aerial robots, some research has been directed towards the development of manipulators for aerial robots in search of more reliable interaction capabilities and strategies. A bird-inspired aerial robotic perching mechanism is presented in \cite{askari2023avian}. Its main innovation is the combination of Hoberman linkage legs and fin ray claws. The weight of the drone is used to wrap the claws around the perch. A passive perching mechanism, capable of perching on complex surfaces and grasping objects, is also presented in \cite{roderick2021bird}, and the important relationship between closed-loop balance control and the range of perching parameters is explored. However, all of these perching mechanisms take the form of perching above tree branches, making it difficult to choose a suitable position when the canopy is dense.

\par In this paper, an aerial robotic perching mechanism based on flexible materials to help drones perch from below the canopy is presented. Firstly, by exploring the friction characteristics of the aerial robot's perch mechanism on a surface, a perching mechanism with a small servo embedded to assist in the perching of aerial robots is designed. Then, a visual segmentation method for target perch selection is designed and evaluated in both simulation and real-world experiments. The experiments confirm that the whole system has the required ability during the perching process. In addition, tendons pulled by the servos will aid in the opening and closing control of the perch mechanism as a way to achieve a stable perch. The whole perching system is shown in \cref{fig1}.
\par The main highlights of this work are as follows:
% \IEEEpubidadjcol

\par1) An aerial perch mechanism based on friction analysis that allows the aerial platform to better adapt to complex surfaces is proposed.
\par2) A perching algorithm with segmentation for branch, which allows for more accurate location selection of target points is proposed, thus aiding the navigation function during drone operations.
\par3) A planning method for the current perching mechanism is implemented so that the trajectory planning and control of the aerial robot can be more compatible with the current mechanism and perching point recognition algorithm.

%%%%%%%%%%%%%%%%%%%%%%%%%%%%%%%%%%%%%%%%%%%%%%%%%%%%
\section{Perching Mechanism Design}
%%%%%%%%%%%%%%%%%%%%%%%%%%%%%%%%%%%%%%%%%%%%%%%%%%%%

\begin{figure}[t]
    \centering
    \includegraphics[width=1\linewidth,trim={100 130 330 50},clip]{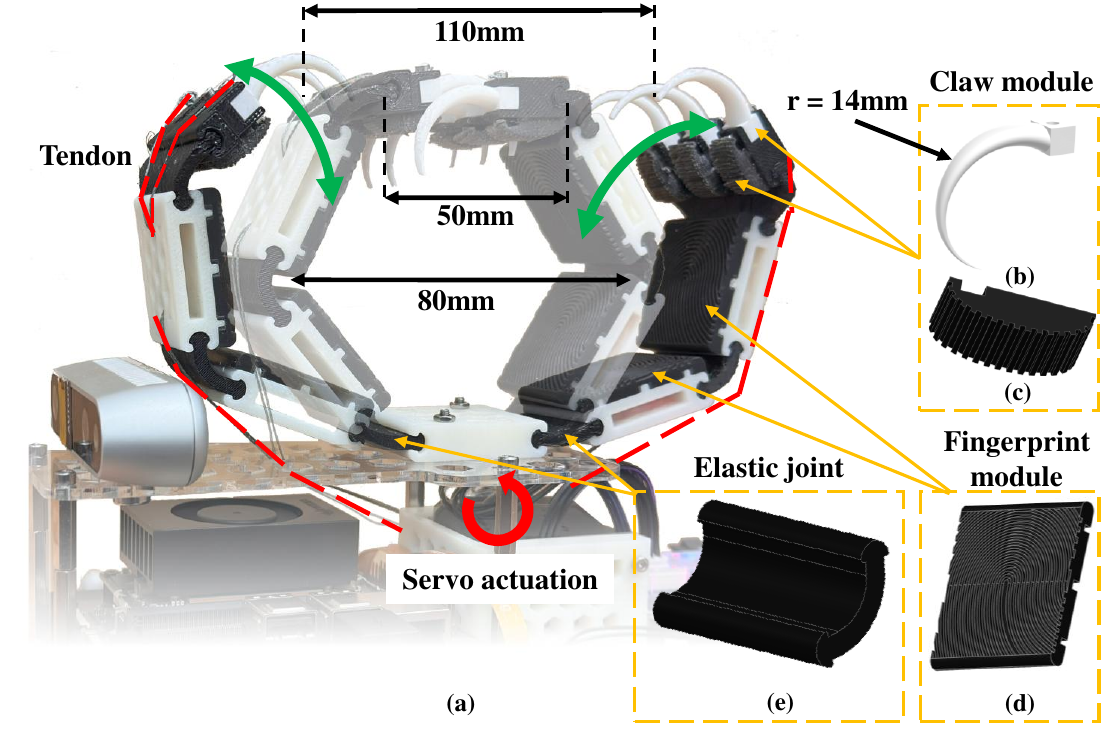}
    \caption{Design and operation of the perching mechanism. (a) The close and open state is achieved through tendon actuation and a servo motor. (b) Claw design. (c) Claw pad for additional friction. (d) Friction pad with fingerprint pattern. (e) Elastic joint.}
    \label{fig:design}
\end{figure}

In this section, a modular design and fabrication scheme is proposed to further improve the utility of the perch mechanism by mixing rigidity and flexibility. Then, the interaction profiles and static models between the clamping mechanism and the target are further analyzed, and based on these, the maximum payload is predicted, which in turn verifies the effectiveness of the actual perching. \cref{fig:design} shows the mechanism and its components.
%%%%%%%%%%%%%%%%%%%%%%%%%%%%%%%%%%%%%%%%%%%%%%%%%%%%
\subsection{Design Conception}
%%%%%%%%%%%%%%%%%%%%%%%%%%%%%%%%%%%%%%%%%%%%%%%%%%%%
To address the problems when performing perching in real environments, careful design of the mechanical element is crucial. It should not only have a reliable payload capacity for the aerial systems but also be energy efficient and maintain stability after the perch without consuming additional energy. It should also adapt irregularly shaped branches with different diameters and surface textures to meet multi-scenario mission requirements.

These requirements are determined by considering the real-world scenarios. During perching in forest environments, possible obstructions such as canopy and leaves can interfere with the flight operation or even damage the platform. To avoid them, the mechanism is designed to be mounted on top of the platform to allow perching from underneath and hanging on the tree branches. Since tree branches vary in diameter, condition and surface texture, compliant and elastic joints are designed to promote perching adaptiveness. These joints are designed to self-maintain the closed state and are actuated via a soft tendon for simple actuation without reducing operation time or adding complexity to control. The mechanism should also be scalable for larger branches, a modular design is therefore adopted to allow an additional range of branch diameter. Inspired by bats whose feet assist them in perching upside down on tree branches, a tendon-driven soft-perching mechanism for aerial platforms is designed. It facilitates both energy efficiency and continued flight out of the perching state.

%%%%%%%%%%%%%%%%%%%%%%%%%%%%%%%%%%%%%%%%%%%%%%%%%%%%
\subsection{Mechanism Design and Fabrication}
% Detailed description of the mechanism: 
%%%%%%%%%%%%%%%%%%%%%%%%%%%%%%%%%%%%%%%%%%%%%%%%%%%%

Guided by the design conception detailed previously, a bio-inspired, compliant, and modularized perching mechanism shown in \cref{fig:design} (a) is designed. The perching mechanism's arm is symmetrical and separated into several modules, which are the claw module that consists of the claw and the finger-pad and the fingerprint module \cite{hao2024friction}, shown in \cref{fig:design} (b), (c) and (d), respectively. The mechanism is actuated by a Dynamixel XL-430 servo which opens the arm by transmitting torque through nylon monofilament as tendons, shown in \cref{fig:design} (a). The closed state is the equilibrium state where no energy is needed to maintain it. To achieve this, the modules are connected by stiff thermoplastic polyurethanes (TPU)-based joints illustrated in Fig. \ref{fig:design} (e) that provide restoration force in the open state and compression force after perching around tree branches. The fingerprint module, shown in \cref{fig:design} (d), can therefore provide extra vertical friction forces to prevent sliding, especially when the mechanism is in full contact with the tree branch. The rigid claw shown in \cref{fig:design} (b) is the end of the mechanism's arm and is designed to provide the majority of the force to hold the overall weight, and is made of a commercially available resin material, Formlabs Rigid 10K, to withstand the impact forces that may be generated by the platform's perching manoeuvre. In \cref{fig:design} (c), a soft, textured finger-pad below the claw fabricated from commercially available photopolymer Agilus 30 can provide extra friction especially when perching on small branches, where fingerprint modules lose contact with the branch surfaces. Thin protrusions are designed to allow interlocking with tree bark irregularities, providing extra resistance force.

%%%%%%%%%%%%%%%%%%%%%%%%%%%%%%%%%%%%%%%%%%%%%%%%%%%%
\subsection{Interaction Profiles and Static Model}
%%%%%%%%%%%%%%%%%%%%%%%%%%%%%%%%%%%%%%%%%%%%%%%%%%%%

The mechanism interacts with the tree branches in three different ways based on branch diameter. In each interaction profile, different modules engage with the branches.

\begin{figure}[t!]
    \centering
\includegraphics[width=1\linewidth]{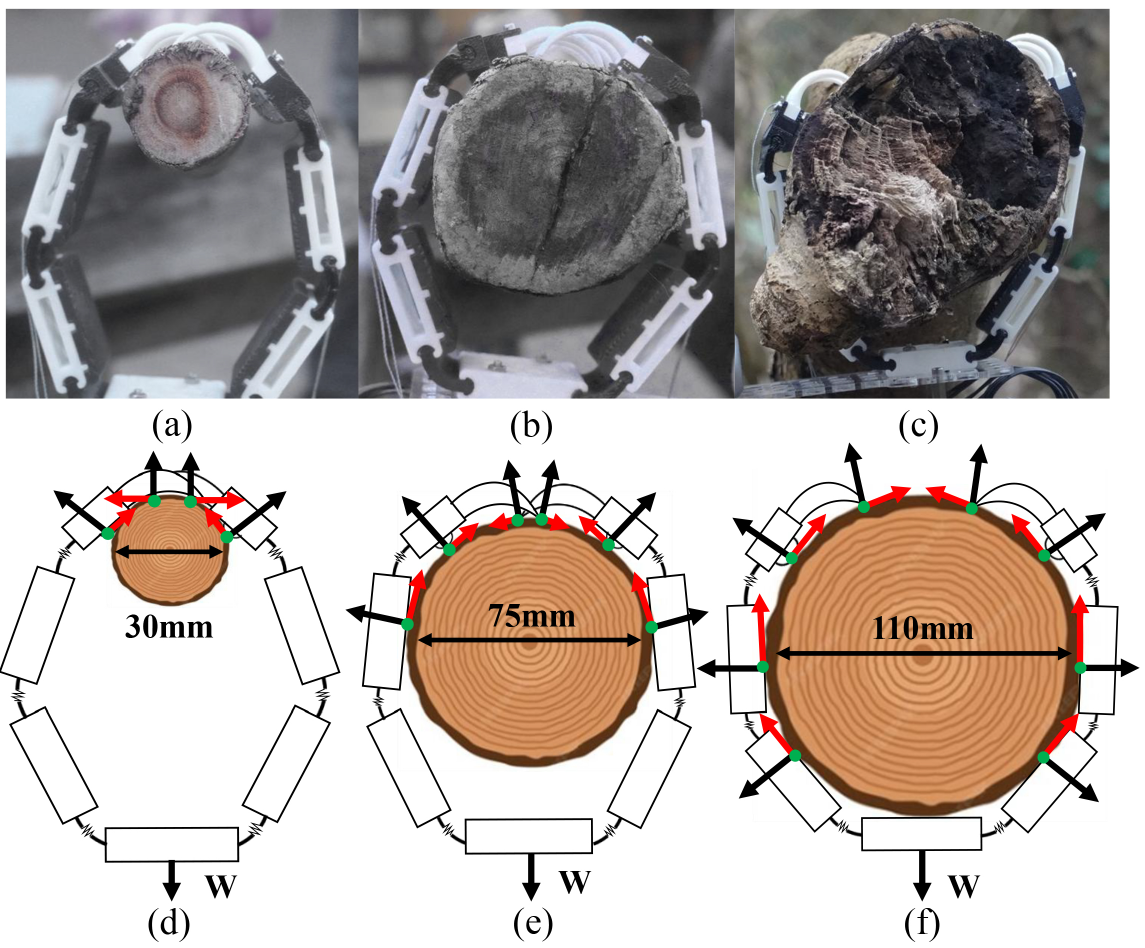}
    \caption{Interaction profile between the branch and the perching mechanism depends on the branch diameter. The green dots show the contact points from an axial view. (a), (d) Small branches. Contact forces are shown in black arrows and the corresponding friction forces are shown in red. (b), (e) Medium branches. Internal joint forces are also shown. (c), (f) Large branches.}
    \label{fig:interaction}
\end{figure}

Depending on the clearances between modules, shown in \cref{fig:design} (a), the three ranges of diameters are: \qtyrange{30}{50}{\milli\metre}, \qtyrange{50}{80}{\milli\metre} and \qtyrange{80}{110}{\milli\metre}. For branches with a diameter less than that of the curvature of the claw, shown in \cref{fig:design} (b), the claws 'hang' on the tree branch, making the other components obsolete. The maximum diameter of the branch can not exceed the range due to the maximum clearance between the claws in the open state, as shown in \cref{fig:design} (a). However, considering the tracking error, the maximum diameter guaranteed by the flying platform is estimated as 80 \si{\milli\metre}. The interactions between the mechanism and the real branches and their corresponding schematic representations performing adaptive manoeuvrers on branches of different diameters are shown in \cref{fig:interaction}.
For branches ranging from \qtyrange{30}{80}{\milli\metre}, the reaction force acting on the claw tip stays relatively aligned with the direction of the weight. In this case, the maximum weight is determined by its material strength and geometry through the curved-beam formula \cite{hibbeler_mechanics_2018} at a location near the tip:

\begin{equation}
F_{\text{max}} = \frac{\sigma_{\text{UTS}}Ar(\bar{r}-R)}{(R-r)L} \label{eq1}
\end{equation}

\noindent where \(\sigma_{\text{UTS}}\) is the ultimate tensile stress of the material, \(A\) is the area of the claw cross-section of which the force, \(R\) is the location of the neutral axis, \(r\) is the distance from the centre of the curvature to the point where the stress is determined, \(\bar{r}\) is the distance from the centre of curvature to the centroid of the cross-section and \(L\) is the moment arm of the predicted force.

\begin{figure}[t]
		\centering
		\begin{overpic}[scale=0.33]{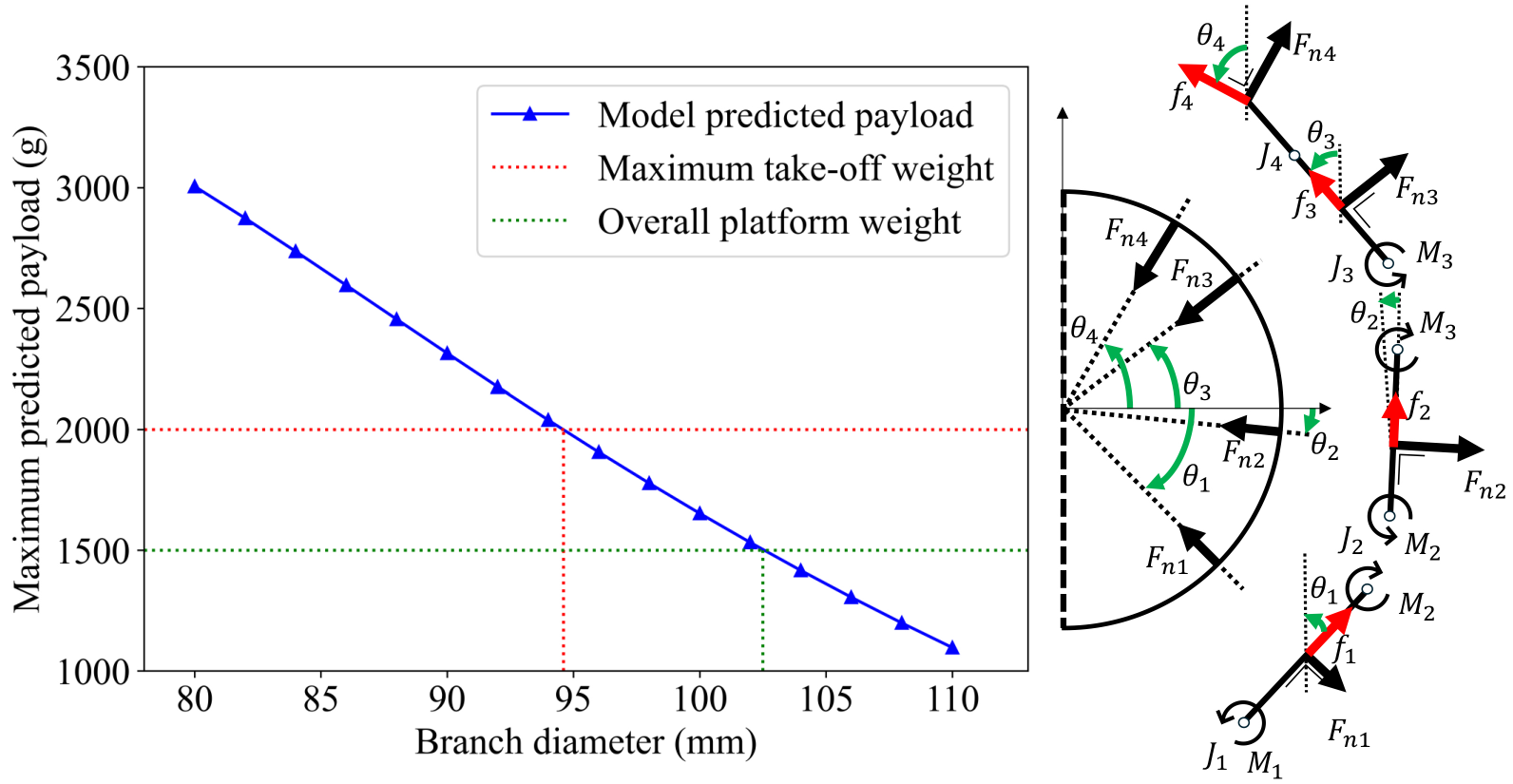}
            \put(36,-2.5){\textcolor{black}{\footnotesize (a)}}
            \put(80,-2.5){\textcolor{black}{\footnotesize (b)}}
            \end{overpic}
            \vspace{-0.4cm}
\caption{Static analysis of the perching mechanism in contact with a target. (a) Predicted payload capacity, the maximum take-off weight is determined by the motor's capacity, while the current platform weighs about 1500 \si{\gram} (b) Simplified static interaction model used for static model.}\label{fig:static}
\end{figure}

 In the case of perching on large branches, the contact happens after the bottom of the mechanism touches the tree branch, resulting in static full contact before perching. During this interaction mode, the 'squeezing force' promoted by the stiff joints becomes more relevant. As the interaction is more complicated, a static model is used to model if the mechanism can withstand the platform weight without slipping. The static model has some key simplifications and assumptions. Firstly, the interaction is simplified as a symmetrical 2D interaction between an ideal circle and a set of straight, connected line segments. The contact point of each module to the tree branch is assumed as the tangent point of the circle and the line segments. A schematic model with relevant forces and joints is shown in \cref{fig:static} (b).

The model begins by solving the geometric problem of finding the angular relationship between connected tangents, which determines the moment that the TPU joints can provide through:
\begin{equation}
M=k\Delta\theta\label{eq}
\end{equation}

\noindent where \(k\) is the stiffness coefficient and \(\Delta\theta\) is the change of angle due to the interaction of each segment. Then working from the top segments, the normal force can be determined via moment equilibrium about the geometrical centre of the joint by:
\begin{equation}
\vec{\bm{M}}_{3}=\vec{\bm{F}}_{n_{4}} \vec{\bm{l}}_{4,3}+
\vec{\bm{F}}_{n_{3}} \vec{\bm{l}}_{3,3}+
\vec{\bm{f}}_{4} \vec{\bm{l}}_{4,3}+
\vec{\bm{f}}_{3} \vec{\bm{l}}_{3,3} \label{eq}
\end{equation}
\begin{equation}
f_4=\mu_{\text{claw}}F_{n_{4}}\label{eq}
\end{equation}
%\begin{equation}
%f_3=\mu_{\text{clawpad}}F_{n_3}\label{eq}
%\end{equation}

\noindent where \(F_{n_{4}}\) denotes the reaction force at the claw tip, \(\vec{\bm{l}}_{4,3}\) is the corresponding vector from the contact point of the claw the joint being considered, where the corresponding moment locates, \(f_{\text{4}}\) is the corresponding friction force and \(\mu_{\text{claw}}\) denotes the coefficient of friction of each contacting component, which is determined through experimenting the friction components with 10 tree-bark samples. By repeating this process over the next two segments, the in-plane reaction force and their corresponding friction force in the vertical direction can be calculated. As shown in \cref{fig:static} (a), when perching on branches with a diameter larger than 102 \si{\milli\metre}, the force is not sufficient to hold the platform's weight. However, in reality, it is noticed that the claw tip can hold onto the tree bark cracks. This transfers the interaction from friction to reaction force subjected to the claw, which can be calculated through \cref{eq1}, providing extra vertical forces other than friction alone. Although the claw is proven to be effective in performing successful perching, the claw-tip will be broken after several engagements, suggesting a strong, less brittle material can be used to support durable operations.

\section{Perching Algorithm Framework} %on the canopy/branch/trunk
In the case of a well-designed perching mechanism, the operational movements of the aerial robot will also affect its perching performance. In order to further improve the perching capability of the proposed aerial platform, the platform's vision, planning, and control algorithms were systematically designed.

%%%%%%%%%%%%%%%%%%%%%%%%%%%%%%%%%%%%%%%%%%%%%%%%%%%%
\subsection{Perching Point Selection}
%%%%%%%%%%%%%%%%%%%%%%%%%%%%%%%%%%%%%%%%%%%%%%%%%%%%

\begin{figure}[h]
		% Requires \usepackage{graphicx} t rim=210 180 230 160,clip,grip
		\centering
		\begin{overpic}[scale=0.5,trim=5 5 5 5,clip]{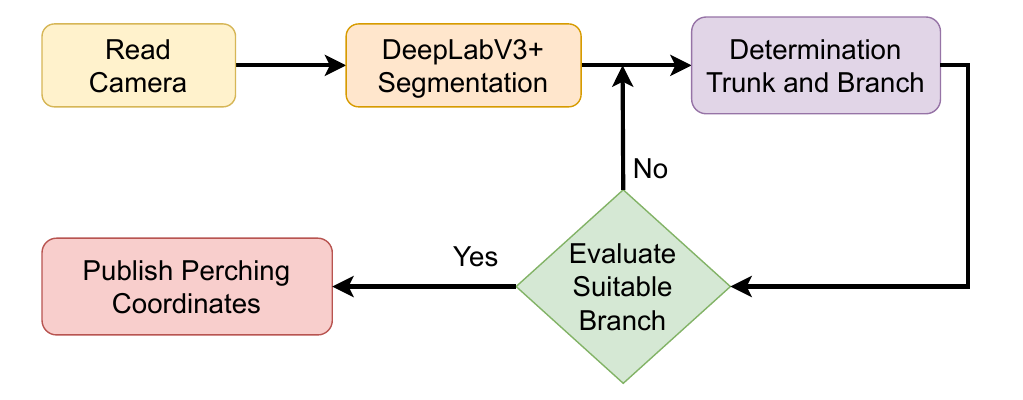} 
		\end{overpic}
\caption{Flow chart of visual segmentation algorithms. Using the visual information from the depth camera as input, the spatial information of the perching point is computed iteratively, and the coordinates of the centre point of the target to be housed are output sequentially. }\label{fig5}
\end{figure}

The proposed perching mechanism already has some tolerance for perching on trunks of different diameters, but for perching tasks in real scenarios, the perch location is also important for systematic perching tasks. To find a suitable perching point, the DeepLabV3+ \cite{deeplabv3_model} architecture is used for semantic image segmentation based on PyTorch for branch and trunk detection, using a ResNet-34 \cite{resnet_34} encoder with pre-trained weights for faster and better convergence. The tree dataset \cite{tree_dataset} with 13 different classes and 1,385 images is used to train a segmentation model. A sigmoid activation function and a learning rate set at 0.001 is adopted. In addition, a dice loss function is created to optimize the performance of the segmentation model using the intersection-over-union (IoU) metric. According to the results obtained, the segmentation performance is calculated to be 72.15\%. The approach used for the decision-making process is shown in \cref{fig5}.

\begin{figure}[b!]
		% Requires \usepackage{graphicx} 
		\centering
		\begin{overpic}[scale=0.29,trim=20 20 100 50,clip]{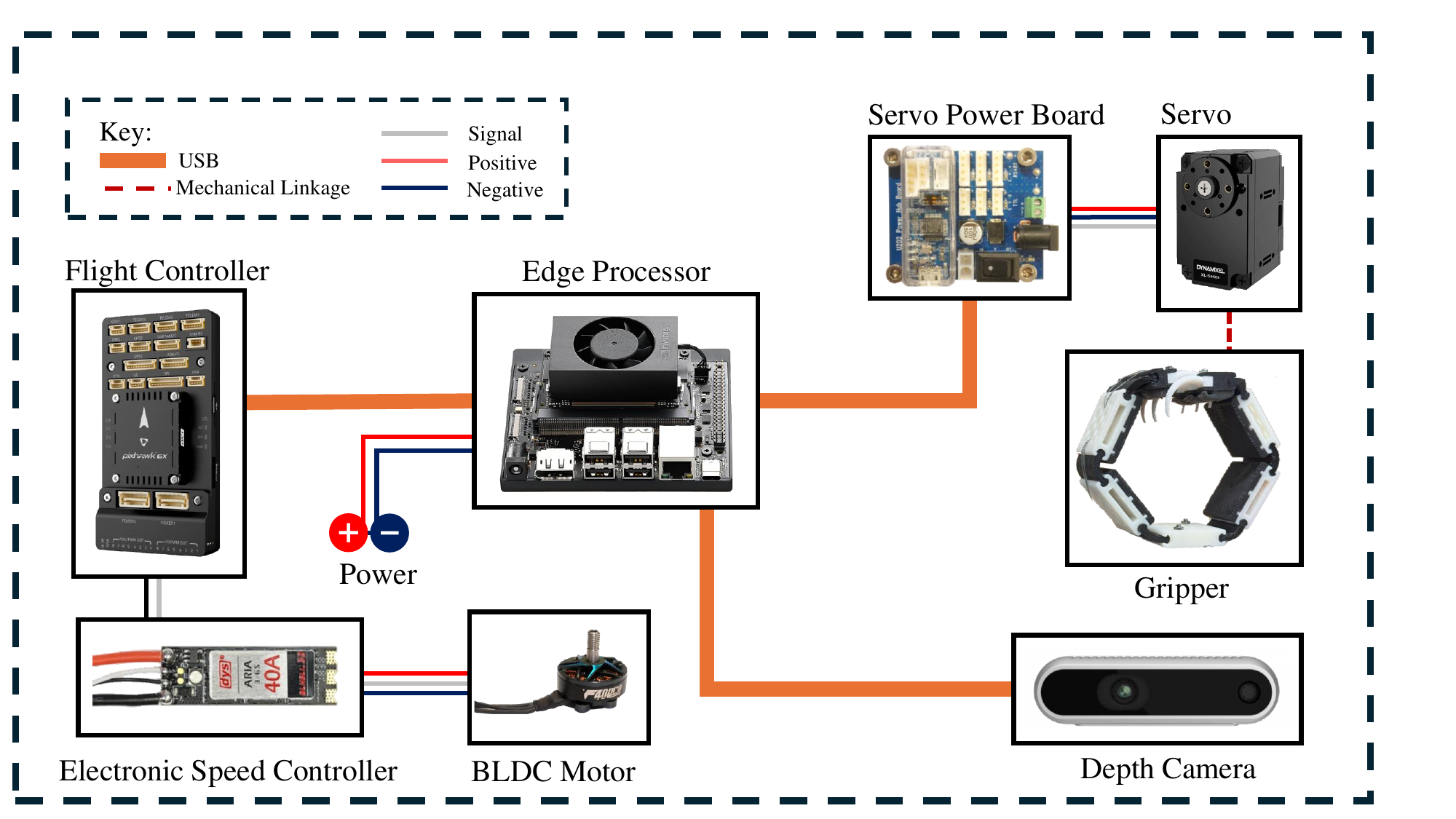} 
		\end{overpic}
\caption{Electromechanical architecture of the aerial perching system. Where the onboard computer receives information from the flight control and depth camera and issues commands to the perching mechanism.}\label{fig6}
\end{figure}

The above approach is inspired by raptors in particular, as they identify the closest trunk to the branch before determining the perching point. It has been observed to perform dynamic perching manoeuvres by first evaluating larger structures for stability and finally fine-tuning their perching on specific branches \cite{raptors_2021}. The vision algorithm first detects the tree trunk and then determines the coordinates of the nearest perch as specified, the planning and control algorithm navigates the aerial robot based on the selected perching point. The main hardware and communication required for the algorithm is shown in \cref{fig6}.
\vspace{-0.2cm}
%%%%%%%%%%%%%%%%%%%%%%%%%%%%%%%%%%%%%%%%%%%%%%%%%%%%
\subsection{Planning and Control}
%%%%%%%%%%%%%%%%%%%%%%%%%%%%%%%%%%%%%%%%%%%%%%%%%%%%
To guide the aerial robotic system to reach the expected position and facilitate the contact of the perching mechanism with the target tree branch, an algorithmic framework was created for the aerial robot to perch on a tree branch, starting from the perch point selection part and continuing to the final stage of perching. In order to make the trajectory of the aerial perching system more dynamically friendly and to ensure the stopping accuracy of the perch position, the minimum-snap\cite{minsnap_2011} trajectory optimization algorithm was integrated into the planning process. 
In the implementation, a 7th-order polynomial to parametrized trajectories is used in each axis. The choice of a 7th-order polynomial ensures sufficient degrees of freedom to enforce continuity up to the third derivative (jerk) at waypoints while minimizing snap. A closed-form solution is used to solve the optimization problem and define the boundary conditions, including position, velocity, acceleration, and jerk at the current state and desired perching point. Then a PID-based linear tracking controller is implemented to track the trajectory.

%%%%%%%%%%%%%%%%%%%%%%%%%%%%%%%%%%%%%%%%%%%%%%%%%%%%
\section{Real-world Perching Experiments}
%%%%%%%%%%%%%%%%%%%%%%%%%%%%%%%%%%%%%%%%%%%%%%%%%%%%

To evaluate the overall performance of the proposed perching mechanism and to further validate the perching capability of the aerial robotic platform, real-world experiments and further analysis of the results are conducted.

%%%%%%%%%%%%%%%%%%%%%%%%%%%%%%%%%%%%%%%%%%%%%%%%%%%%
\subsection{Experimental Setup}
%%%%%%%%%%%%%%%%%%%%%%%%%%%%%%%%%%%%%%%%%%%%%%%%%%%%

A dummy tree was built for the experiment,  consisting of a 1.27 \si{\metre} long branch with a diameter of 30 \si{\milli\metre},  placed at a height of 1.8 \si{\metre}. The perching mechanism is a combination of a main body made of polylactic acid material and a non-slip surface made of thermoplastic polyurethane material (TPU), and was integrated into a quadcopter for aerial perching tasks. An NVIDIA Jetson Orin Nano was chosen as the mission computer to communicate with the flight controller and Intel RealSense D435i camera through its serial interface while running the target extraction and mission-related algorithms. The Holybro Pixhawk 6X flight controller was chosen to control the robot’s flight and the servo control for the perch mechanism reset. The VICON motion capture system was used for state estimation with using an extended Kalman filter to fuse state estimation with an onboard inertia measurement unit.

%%%%%%%%%%%%%%%%%%%%%%%%%%%%%%%%%%%%%%%%%%%%%%%%%%%%
\subsection{Perching Point Selection}
%%%%%%%%%%%%%%%%%%%%%%%%%%%%%%%%%%%%%%%%%%%%%%%%%%%%

\begin{figure}[t!]
		% Requires \usepackage{graphicx} trim=210 180 230 160,clip,grip
		\centering
		\begin{overpic}[scale=0.35]{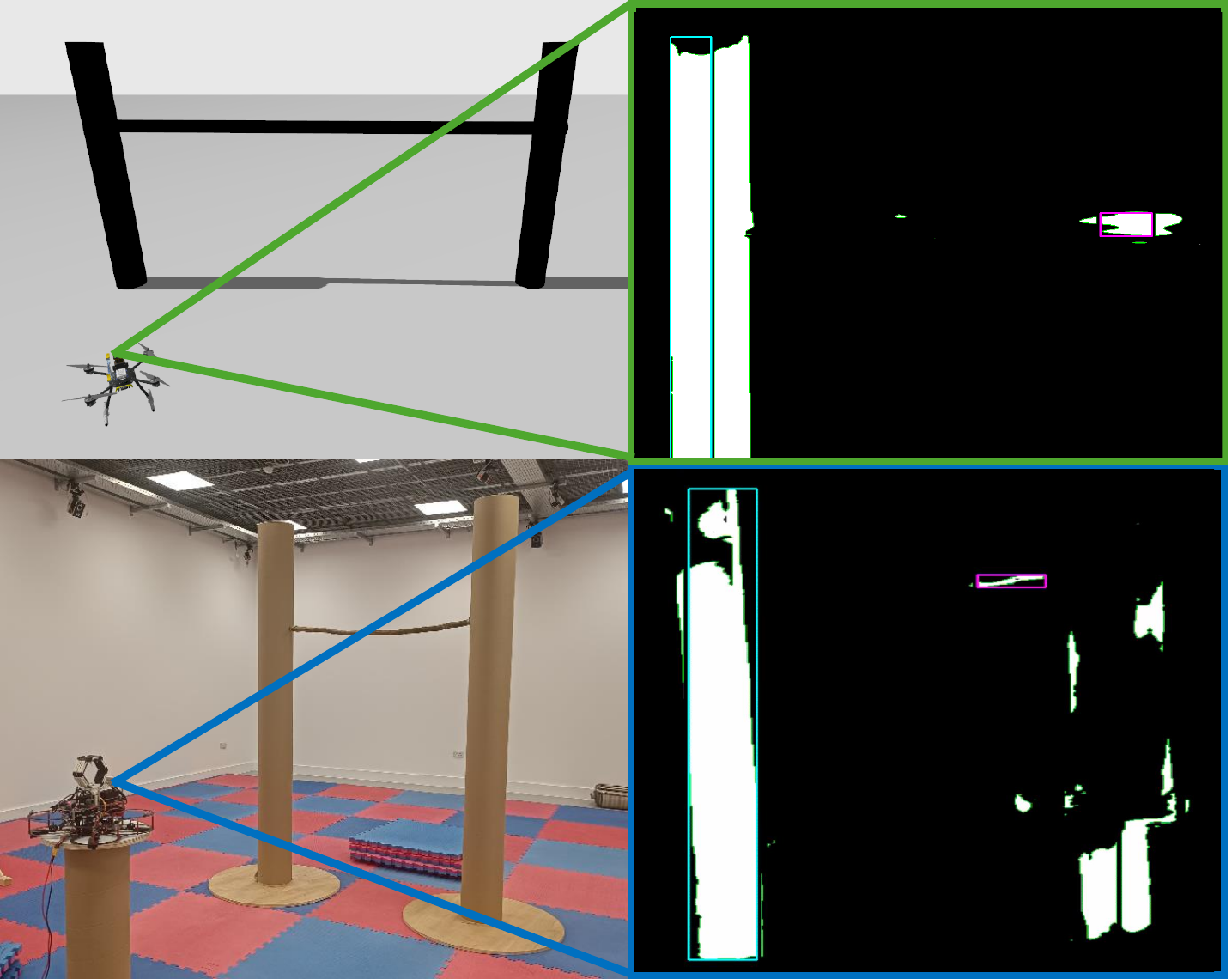} 
            \put(0.5,76){\textcolor{black}{\textbf{(a)}}}
            \put(0.5,37){\textcolor{black}{\textbf{(b)}}}
		\end{overpic}
\caption{ Illustration of the perching point selection. The pink and cyan bounding boxes represent perching point selection and trunk prediction, respectively. \textbf{(a)} shows the result of the algorithm in the simulated environment and \textbf{(b)} shows the result of the algorithm in the real world.}\label{fig7}
\end{figure}

The algorithms of the perching point selection were evaluated in simulation and real-world scenarios, respectively. In the Gazebo-based simulation environment, the target to be perched was similar to that in the real environment and consisted of a cylinder with the same dimensions as the experimental setup described above. A comparison of the environment configuration in the real world and the simulation environment as well as the effect of the visual algorithm is visualised in \cref{fig7}, where the triangular boxes are the schematic of the depth camera's field-of-view, and the cyan and pink contours in the depth map on the right are the recognized tree trunks and branches, respectively.

\begin{figure}[b!]
		% Requires \usepackage{graphicx} trim=210 180 230 160,clip,grip
		\centering
		\begin{overpic}[scale=0.45,trim=5 5 5 5,clip]{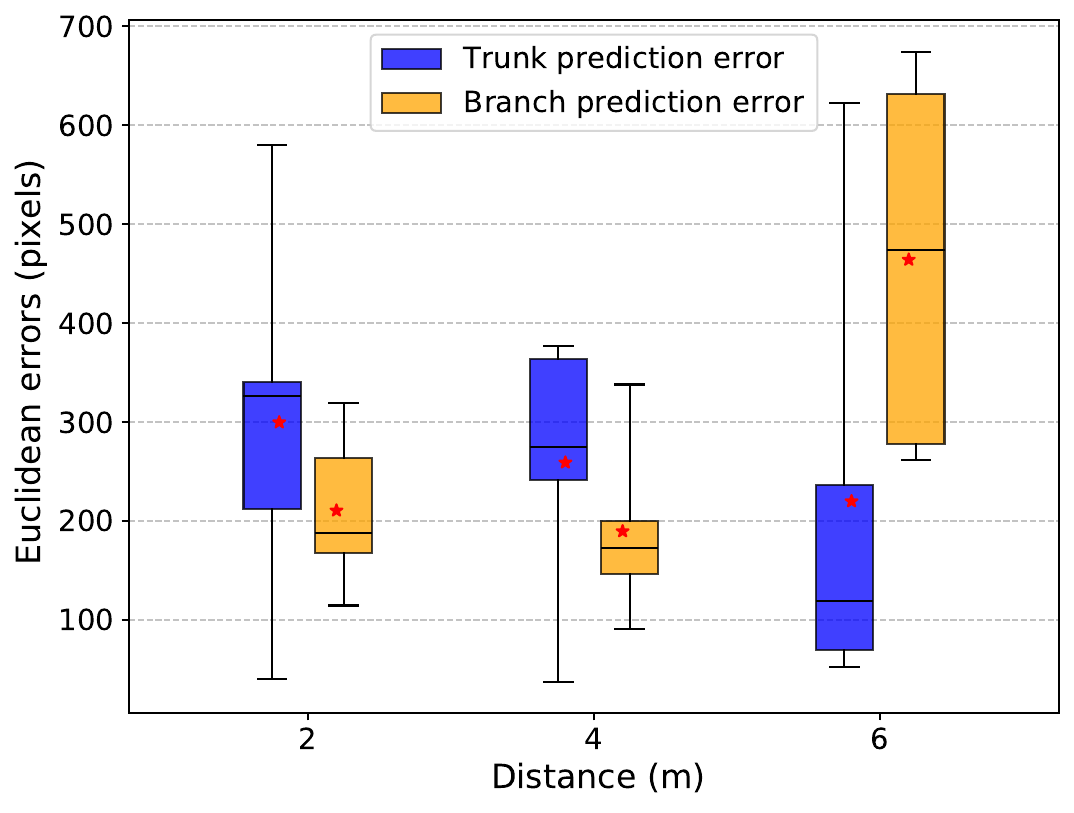}            
		\end{overpic}
\caption{ Euclidean error and standard deviation of trunk and branch predictions at different distances in the Gazebo environment. On a box plot, the horizontal line indicates the median and the red '*' symbol indicates the mean. The bottom and top edges of the box illustrate the 25th and 75th percentiles, respectively. }\label{fig8}
\end{figure}

In both simulation and real-world scenarios, a perching point selection algorithm was implemented, which first detected a trunk and then found a suitable branch closest to the trunk to perch on. The results based on simulations and the real-world environment show that the proposed algorithm can separate the trunk from the branches and use the centre coordinate point of the suitable branch as an output to be used for perching. To evaluate the algorithm used for perching point and trunk detection, tests from three different distances were conducted. In the Gazebo simulation environment, the trunk detection and perching point selection performance of the visual segmentation algorithm was evaluated according to the Euclidean error criterion. The results of a total of 15 different runs at three distances are shown in \cref{fig8}.

The results show that the branch prediction error increases significantly after a distance of 4 \si{\metre}. However, the trunk prediction error remains within a certain range. It can be concluded from the results that 4 \si{\metre} is more suitable than other distances for both trunk and branch detection. Because when the platform is 1 \si{\metre} away to the trunk, only sections of the trunk can be viewed by the camera, and when the platform is 6 \si{\metre} away, detecting the branch becomes difficult as the target is too small in the camera frame. Therefore, it is more feasible to detect the trunk and the perching point at a distance around 4 \si{\metre}.
%%%%%%%%%%%%%%%%%%%%%%%%%%%%%%%%%%%%%%%%%%%%%%%%%%%%
\subsection{Planning and Control}
%%%%%%%%%%%%%%%%%%%%%%%%%%%%%%%%%%%%%%%%%%%%%%%%%%%%

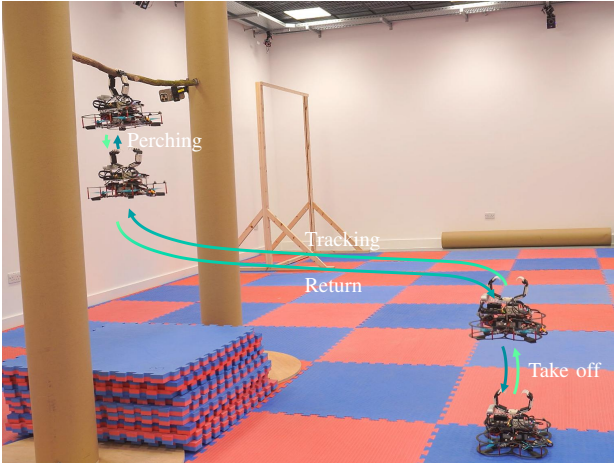
\begin{figure}[t!]
		% Requires \usepackage{graphicx} trim=210 180 230 160,clip,grip
		\centering
		\begin{overpic}[scale=0.37,trim=140 30 150 40,clip]{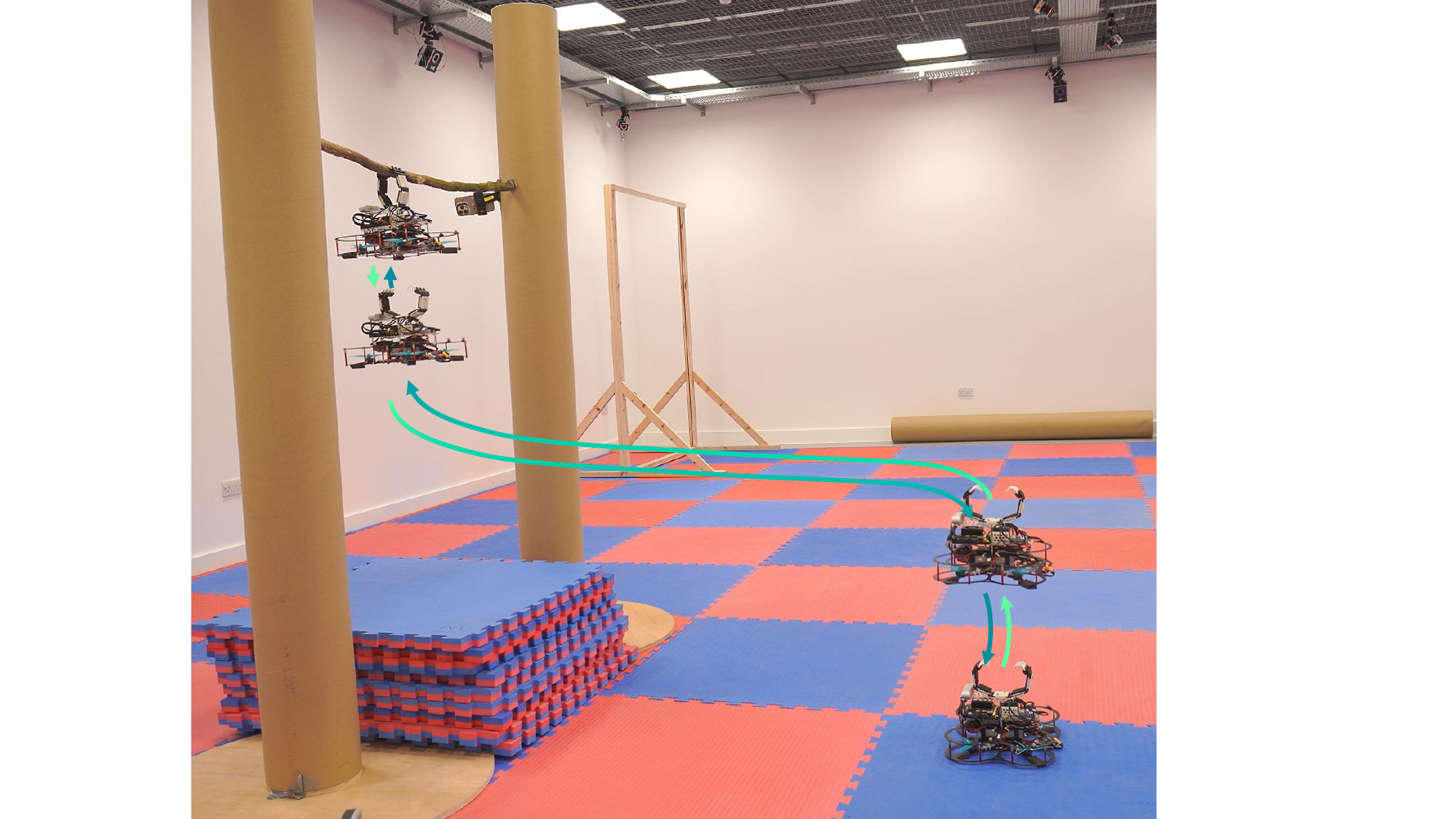} 
            \put(80,13){\textcolor{white}{\footnotesize Take off}}
            \put(46,33){\textcolor{white}{\footnotesize Tracking}}
            \put(46,26){\textcolor{white}{\footnotesize Return}}
            \put(19,48){\textcolor{white}{\footnotesize Perching}}
		\end{overpic}
\caption{Illustration of the real-world perching experiment. The aerial robotic system starts with takeoff, then tracks the trajectory, and finally executes the perching manoeuvre.}\label{fig9}
\end{figure}

The experimental procedure for a real-world perching mission includes five main stages: take-off, trajectory planning, trajectory tracking, perching and resume. First, the quadrotor takes off to a predefined hovering position. Next, the minimum-snap planning method plans a trajectory from the hover position to the selected perching point. The planned trajectory is then executed and navigated by the quadrotor along the trajectory with the linear controller. Upon reaching the final waypoint near the selected perching point, the gripper mechanism is triggered to grasp the branch. Finally, the flight controller is disarmed to transition the quadrotor into perched mode. The real-world experimental results show that the trajectory planning and control method is effective for perching tasks. The image sequence of a real experiment is visualised in \cref{fig9}.

% The aerial robot demonstrated smooth and accurate navigation by closely following the planned trajectory in the X, Y and Z axes. The effect in the real experiment is shown in \cref{fig9}.

The experimental trajectory tracking of default PID controller results indicated a mean 3D position tracking error of $0.122$ \si{\metre} with a standard deviation of $0.054$ \si{\metre}, reflecting the accuracy of the trajectory tracking controller. Analysing each axes, the X-axis exhibited the mean error at $0.029$ \si{\metre} with a standard deviation of $0.018$ \si{\metre}. The Y-axis showed a mean error of $0.075$ \si{\metre} with a standard deviation of $0.023$ \si{\metre}. Similarly, the Z-axis reported a mean error of $0.079$ \si{\metre} with a standard deviation of $0.066$ \si{\metre}. The error analysis is organized as shown in Table \ref{tab1}. It is expected that this error could be compensated by the mechanical structure of the mechanism and additional tuning of the PID controller.

\begin{table}[b!]
\centering
\caption{Position Tracking Error.}
\label{tab:my-table}
% \resizebox{\columnwidth}{!}{%
\begin{tabular}{ccc}
\toprule
\textbf{Axis}        & \multicolumn{2}{c}{\textbf{Position Tracking Error}} \\
                     & \textbf{Mean {[}m{]}}     & \textbf{STD {[}m{]}}     \\ \midrule
\rowcolor[HTML]{F1F1F2} 
X                    & 0.029                     & 0.018                    \\
Y                    & 0.075                     & 0.023                    \\
\rowcolor[HTML]{F1F1F2} 
Z                    & 0.079                     & 0.066                    \\ \midrule
\textbf{3D Position} & \textbf{0.122}            & \textbf{0.054}           \\ \bottomrule
\end{tabular} \label{tab1}
% }
\end{table}

The current work has limitations with a simplified computer vision formulation for a clean background and an insufficient number of real experimental evaluations to claim statistical significance, and they will be part of the continuation of future work.

%%%%%%%%%%%%%%%%%%%%%%%%%%%%%%%%%%%%%%%%%%%%%%%%%%%%
\section{Conclusion}
%%%%%%%%%%%%%%%%%%%%%%%%%%%%%%%%%%%%%%%%%%%%%%%%%%%%

In this paper, a mechanical design has been presented for an aerial robotic system that can perch on suitable branches in the air based on visual segmentation. An adaptive gripping mechanism controlled by a servo and tendons is first designed. The mechanism is proved suitable for perching on branches with a range of diameter from 30 \si{\milli \metre} to 80 \si{\milli \metre}, without having to consider shape, condition or surface texture of the branch. Then, to find a perching position, we design a visual segmentation algorithm for outputting 3D coordinate points in a suitable position. Finally, the trajectory of the aerial robot was planned and controlled to accomplish perching manoeuvres on aerial branches.
\par Initial experimental evaluations show that our aerial robotic platform is promising in acquiring the centre coordinates of the target to be perched and can perch well on tree trunks of different diameters in the tolerance of the proposed perching mechanism. 
Future work would also need to quantify the success rates in multiple starting points with multiple sets of experiments to release the bottlenecks of the proposed framework systematically.

\section*{Acknowledgments}
% This should be a simple paragraph before the References to thank those individuals and institutions who have supported your work on this article.
We gratefully acknowledge the Bristol Robotics Laboratory for providing access to the flight arena. This research was also supported by seedcorn funds by Civil, Aerospace and Design Engineering, Isambard AI and Bristol Digital Futures Institute at the University of Bristol (Worktribe ID: 2695947). We also extend our appreciation to technician Patrick Brinson for his valuable assistance in conducting the experiments. Furthermore, we acknowledge Chuanbeibei Shi and Halim Atli for their initial explorations with this project's concept, and Alex Dunnett for his contributions to photo editing and proofreading of this work.

% {\appendix[Proof of the Zonklar Equations]
% Use $\backslash${\tt{appendix}} if you have a single appendix:
% Do not use $\backslash${\tt{section}} anymore after $\backslash${\tt{appendix}}, only $\backslash${\tt{section*}}.
% If you have multiple appendixes use $\backslash${\tt{appendices}} then use $\backslash${\tt{section}} to start each appendix.
% You must declare a $\backslash${\tt{section}} before using any $\backslash${\tt{subsection}} or using $\backslash${\tt{label}} ($\backslash${\tt{appendices}} by itself
%  starts a section numbered zero.)}

% Space
\addtolength{\textheight}{-12cm}

%{\appendices
%\section*{Proof of the First Zonklar Equation}
%Appendix one text goes here.
% You can choose not to have a title for an appendix if you want by leaving the argument blank
%\section*{Proof of the Second Zonklar Equation}
%Appendix two text goes here.}

% \bibliography{ref.bib}{}
% \bibliographystyle{unsrt}
\bibliographystyle{IEEEtran}
\bibliography{IEEEabrv,ref.bib}

\end{document}